\documentclass[runningheads,a4paper]{llncs}

\usepackage{amssymb}
\usepackage{graphicx}
\usepackage{epstopdf}
\usepackage{color}
\usepackage{url}
\usepackage{epstopdf}
\usepackage{listings}
\usepackage[square,numbers]{natbib}
\usepackage{tabu}
\usepackage[table]{xcolor}
\usepackage{enumitem}
\usepackage{hyphenat}
\urldef{\maill3s}\path|{holzmann, risse}@L3S.de|    
\newcommand{\keywords}[1]{\par\addvspace\baselineskip
\noindent\keywordname\enspace\ignorespaces#1}

\begin{document}

\mainmatter

\title{Insights into Entity Name Evolution\newline{}on Wikipedia\thanks{This work is partly funded by the European Research Council under ALEXANDRIA (ERC 339233)}}

\titlerunning{Insights into Entity Name Evolution on Wikipedia}

\author{Helge Holzmann
\and Thomas Risse}
\authorrunning{Holzmann, Risse}

\institute{L3S Research Center\\
Appelstr. 9a, 30167 Hanover, Germany\\
\maill3s\\
\url{http://www.L3S.de}}

\maketitle

\begin{abstract}
Working with Web archives raises a number of issues caused by their temporal characteristics. Depending on the age of the content, additional knowledge might be needed to find and understand older texts. Especially facts about entities are subject to change. Most severe in terms of information retrieval are name changes. In order to find entities that have changed their name over time, search engines need to be aware of this evolution. We tackle this problem by analyzing Wikipedia in terms of entity evolutions mentioned in articles regardless the structural elements. We gathered statistics and automatically extracted minimum excerpts covering name changes by incorporating lists dedicated to that subject. In future work, these excerpts are going to be used to discover patterns and detect changes in other sources. In this work we investigate whether or not Wikipedia is a suitable source for extracting the required knowledge.
\end{abstract}



\section{Introduction}

Archiving is important for preserving knowledge. With the Web becoming a core part of the daily life, also the archiving of Web content is gaining more interest. However, while the current Web has become easily accessible through modern search engines, accessing Web archives in a similar way raises new challenges. Historical and evolution knowledge is essential to understand archived texts that were created a longer time ago. This is particularly important for finding entities that have evolved over time, such as the US president Barack Obama, the state of Czech Republic, the Finnish phone company Nokia, the city of Saint Petersburg or even the Internet. These entities have not always been as we know them today. There are several characteristics that have changed: their jobs, their roles, their locations, their governments, their relations and even their names. For instance, did you know an early version of the Internet was called Arpanet? Maybe you know, but are future generation going to know this? And how can they find information about the early Web without knowing the right terms? A name is the most essential parameter for information retrieval regarding an entity. Therefore, we need to know what the entity used to be named in former days. Or even more convenient, search engines should directly take it into account.

Similar to search engines, many data centric applications rely on knowledge from external sources. Semantic Web knowledge bases are built upon ontologies, which are linked among each other. These \textit{Linked Data Sources} can be conceived as the mind of the modern Web. However, other than human minds, it represents just the current state, because none of the popular knowledge bases, like DBPedia\cite{DBpedia}, Freebase\cite{Freebase}, WordNet\cite{Wordnet} or Yago~\cite{YAGO}, provide explicit evolution information. Even though some of them include alternative names of entities, they do not provide further details, e.g. whether or not a name is still valid and when a name was introduced.

Although much of this data has been extracted from Wikipedia and Wikipedia provides evolution information to a certain extent, it is missing in these knowledge bases. The reasons are obvious: Wikipedia provides current facts of an entity in semi-structured info boxes, which are simple to parse and therefore often used for gathering data. Evolution data on the other hand is mostly hidden in the texts. 

In this paper we will show how to collect this information by incorporating lists that are available on Wikipedia and provide name evolution information in a semi-structured form (s. Section~\ref{sec:analysis}). We wanted to understand whether or not Wikipedia can be used as a resource for former names of entities. Eventually, we will show that it provides a good starting point for discovering evolutions. In future work, these findings can be used for pattern extraction. The patterns in turn can help to discover more names, even in different sources. Using the data from Wikipedia, we might be able to do this in future work in a completely automatic way. By analyzing the structures of Wikipedia, we obtained some interesting insights on how name changes are handled (s. Section~\ref{sec:name_changes_on_Wikipedia}). The main contributions of this work are:

\begin{itemize}
\item We present detailed statistics on entity name changes in Wiki\-pedia's content beyond structural meta data (s. Section~\ref{sec:lists-entity-name}).
\item We show that a large majority of more than two-thirds of the name changes mentioned in Wikipedia articles occur in excerpts in a distance of less than three sentences (s. Section~\ref{sec:ment-wikip-articl}).
\item We detect and extract those excerpts without manual effort by incorporating dedicated lists of name changes on Wiki\-pedia (s. Section~\ref{sec:procedure}).
\end{itemize}

\section{Handling of Name Changes on Wikipedia}
\label{sec:name_changes_on_Wikipedia}

Wikipedia does not provide a functionality to preserve historic data explicitly. There is no feature to describe the evolution of an entity in a structured way. The only historic information that is made available is the revision history. However, this information does not describe the evolution of an entity, it rather shows the development of the corresponding article. These two things may correspond but they do not have to. There are a number of reasons for a change of a Wikipedia page, even though the actual name of the corresponding article did not change. One of the reasons is that an article did not follow Wikipedia's naming conventions 
in the first place and had to be corrected. By exploiting actual revised versions of an article rather than its history, \citet{Kanhabua2010} achieved promising results in discovering former names from anchor texts. However, in terms of the time of evolution, the only information available is the revision date of the article, as stated above.

According to Wikipedia's guidelines, the title of an article should correspond to the current name of the described entity: ``\textit{if the subject of an article changes its name, it is reasonable to consider the usage since the change}''\footnote{\url{http://en.wikipedia.org/wiki/Wikipedia:Naming_conventions\#Use_commonly_recognizable_names}}. If the official name of an entity changes, the article's name should be changed as well and a redirect from the former name should be created\footnote{\url{http://en.wikipedia.org/wiki/Wikipedia:Move}}. Hence, the only trace of the evolution is the redirection from the former name. However, important details like the information that this has been done as consequence of an official name change are missing.

Another guideline of Wikipedia advices to mention previous and historic names of an entity in the article\footnote{\url{http://en.wikipedia.org/wiki/Wikipedia:Official_names\#Where_there_is_an_official_name_that_is_not_the_article_title}}: ``\textit{Disputed, previous or historic official names should [\ldots be \ldots] similarly treated in the article introduction in most cases. [\ldots] The alternative name should be mentioned early (normally in the first sentence) in an appropriate section of the article.}''. Accordingly, the text is likely to contain name evolution information in an unstructured form, which is hard to extract by automatic methods. An approach to tackle this would be recognizing patterns like \textit{[former name] ... has changed to [new name], because of ... in [year]}. We will investigate if name evolutions are described in such short excerpts of limited length, which are dedicated to the changes, rather than spread through the article. 


To explore these excerpts that describe name evolutions we utilized list pages on Wikipedia. These are specialized pages dedicated to list a collection of common data, such as entity name changes, in a semi-structured way. An example of such a list page is the \textit{List of city name changes}\footnote{\url{http://en.wikipedia.org/wiki/List_of_city_name_changes}}. Name changes on this list are presented as arrow-separated names, e.g. \textit{Edo $\rightarrow$ Tokyo (1868)}.
Unfortunately not all lists on Wikipedia have the same format. While the format of this list is easy to parse, others, such as the \textit{List of renamed products}\footnote{\url{http://en.wikipedia.org/wiki/List_of_renamed_products}}, consist of prose describing the changes, like: ``\textit{Dime Bar, a confectionery product from Kraft Foods was rebranded Daim bar in the United Kingdom in September 2005 [\ldots].}''

As the lists are not generated automatically, also the presentation of the items within a single list can vary. For instance, while some contain a year, other do not, but include additional information instead. Another drawback of the lists is that they are not bound to the data of the corresponding articles. Therefore, a name change that is covered by such a list is not necessarily mentioned in the article, neither the other way around. By analyzing the coverage of name changes in Wikipedia articles more deeply and gathering statistics by focusing on the content rather than Wikipedia's structure, we wanted to answer the question: \textit{Is Wikipedia a suitable seed for automatically discovering name evolutions in texts?}

\section{Analysis Method}
\label{sec:analysis}

In the following we define our research questions and describe the procedure we followed during gathering the statistics.

\subsection{Research Questions}
\label{sec:research_questions}

In our study we want to investigate to what extent Wikipedia provides information about name evolutions and the way these are presented. The goal is to reinforce our hypothesis, that name evolutions are described by short dedicated excerpts (compare Section~\ref{sec:name_changes_on_Wikipedia}). Based on this hypothesis, we derived more specific research question, which will be answered in following sections.

We started our analysis with the question: \textit{How are name evolutions handled and mentioned on Wikipedia?} This question has already been partially answered in Section~\ref{sec:name_changes_on_Wikipedia}, based on the structure and guidelines of Wikipedia. As described, name evolutions are not available in a fully structured form, but rather mentioned in prose within the articles of the corresponding entities. Additionally, some name changes are available in a semi-structured manner on specialized list pages, which serve as a starting point for our research. Ultimately, our goal is to use the data from Wikipedia to build up a knowledge base dedicated to name evolution. 
In order to investigate if Wikipedia and the available lists of name changes can serve a seed for extracting suitable patterns, we require statistical data. Most important is the question: \textit{How many complete name changes, consisting of preceding name, succeeding name and change date, are mentioned in these articles?}. 

With the curiosity provoked by these statistics we extracted minimum parts from the articles that include the names of a change together with the date. In future work, we want to use these texts to extract patterns for discovering name evolutions in other Wikipedia articles or texts from other sources. This raised the question: \textit{Do pieces of texts of limited length exist that are dedicated to a name change?}. The availability of a limited length is required for the extraction task. Otherwise, a pattern-based approach would not be applicable and this task would become more complex.

The final question we want to answer is: \textit{How many sentences do excerpts span that mention a complete name change?} We will analyze percentages of different lengths in order find a reasonable number of sentences for the extraction in other articles later on. This is going to help us validating the initial hypothesis.

\begin{lstlisting}[frame=single,language=Ruby,captionpos=b,numbers=none,float=t,caption={Algorithm in Ruby for computing the sentence distance of the minimum excerpt covering a name change.},escapechar=@,columns=flexible,keepspaces=true,label=lst:distance_algorithm]
 # computing min_distance of name change "P -> S (D)"
 sentences  = [@\textit{pseudo: Array with extracted sentences}@]
 preceding  = [@\textit{pseudo: Array with indices of sentences containing P}@].sort
 succeeding = [@\textit{pseudo: Array with indices of sentences containing S}@].sort
 date       = [@\textit{preudo: Array with indices of sentences containing D}@].sort
 components = [preceding, succeeding, date]
 min_distance, min_from, min_to = nil, nil, nil
 # until no index is available for one component
 # (components.first is on that hass been previously shifted)
 until components.first.empty?
   components.sort! # sort P, S, D by their next index
   from     = components.first.shift # remove smallest
   to       = components.last.first
   distance = to - from
   # save excerpts bounds for new min_distance
   if min_distance.nil? || distance < min_distance
     min_distance, min_from, min_to = distance, from, to
   end
 end
 # min excerpt spanning from sentence min_from to min_to
 exists = !min_distance.nil?
 excerpt = sentences[min_from..min_to].join(' ') if exists
\end{lstlisting}

\subsection{Procedure}
\label{sec:procedure}

The analysis was conducted in two steps. First, we created the dataset as described in Section~\ref{sec:dataset}. Afterwards, the actual analysis was perfomed by collecting and analyzing the statistical information. 


\subsubsection{Gathering Data}
\label{sec:gathering_data}

The lists of name changes from Wikipedia constitute our set of name evolutions. Each item in these lists represents one entity and has the format \textit{Former Name $\rightarrow$ ... $\rightarrow$ Preceding Name $\rightarrow$ Succeeding Name $\rightarrow$ ... $\rightarrow$ Current Name}. Every pair of names delimited by an arrow is considered as a name change. Some of the names are followed by parenthesis containing additional data. Mostly, these are the dates of the change. However, there are cases where the brackets contain alternative names, spelling variations (e.g., the name in another language) or additional prose, like ``changed'', which is redundant in this context and means noise for our purpose. As we are only interested in the dates as well as name variations, we used the brackets containing numbers to extract dates and from the remaining we collected those starting with a capital letter as aliases.

While parsing the lists, we downloaded the corresponding Wikipedia article for each name of an entity by following the links on the names. In case no links were available, we fetched the article with the name or one of the aliases, whatever could be resolved first. For entities with names referring to different articles, we downloaded all of them. If the article redirected to another article on Wikipedia, the redirection was followed. During extraction all HTML tags were removed and we only kept the main article text.

\subsubsection{Analysis}
In the second step, we gathered statistics from the extracted data. The results are explained in the next section. Since one of the objectives of this analysis is to identify the number of sentences spanned by excerpts that cover a name evolution (s. Section~\ref{sec:research_questions}), we split the articles into sentences. This was done by using the sentence split feature of Stanford's CoreNLP suite \cite{CoreNLP}.

Entity names were identified within sentences by performing a case insensitive string comparison. We required either a match of the name itself or one of its aliases, as extracted from the lists, to consider the name mentioned in a sentence. Due to the different formats that dates can be expressed in, we only matched the year of a date in a sentence.

Finally to analyze the sentence distances of all three components of a name change the algorithm shown in Listing~\ref{lst:distance_algorithm} has been used. It computes the minimum distance between the first sentence and the last sentence of an excerpt that includes a complete name change within a text. As an example consider the following text about Swindon, a town in South West England, taken from the Wikipedia article of Swindon\footnote{\url{http://en.wikipedia.org/wiki/Swindon}}.

``\textit{On 1 April \textbf{1997} it was made administratively independent of Wiltshire County Council, with its council becoming a new unitary authority. It adopted the name \textbf{Swindon} on 24 April \textbf{1997}. The former \textbf{Thamesdown} name and logo are still used by the main local bus company of \textbf{Swindon}, called \textbf{Thamesdown} Transport Limited.}''

It describes the name change from Swindon's former name Thamesdown to its current name in 1997, which is mentioned on the Wikipedia page of geographical renaming\footnote{\url{http://en.wikipedia.org/wiki/Geographical_renaming}} as \textit{Thamesdown $\rightarrow$ Swindon (1997)}. The excerpt spans three sentences and includes all three components of the name change twice: Swindon is contained in the sentences with index 1 and 2, Thamesdown is contained twice in sentence 2 and the change date (1997) is mentioned in the sentences 0 and 1. Therefore, one possible sentence distance would be 2, spanning the entire excerpt. However, without the first sentence still all components are included with a shorter distance of 1 (from sentence 1 to sentence 2). This is the minimum sentence distance of this change. If all three components are included in one sentence, the distance would be 0.

\section{Analysis Results}
\label{sec:results}

In this section we present the results of our analysis, based on the research questions from Section~\ref{sec:research_questions}. After describing the used Wikipedia dataset, we present statistical results and observations we made during the analysis and final discussions.

\subsection{Wikipedia Dataset}
\label{sec:dataset}

The data we used for our analysis was collected from the English Wikipedia on February 13, 2014. As starting point we used list pages dealing with name changes. Due to the different formats of these lists
we focused on lists of the form that each entity is represented as a single bullet and the name changes are depicted by arrows ($\rightarrow$) (s. Section~\ref{sec:name_changes_on_Wikipedia}). These lists are easy to parse and therefore, provide a reliable foundation in terms of parsing errors. We found 19 of those lists on Wikipedia. Nine of the lists were identified as fully redundant. After filtering redundant items, we ended up with 10 lists: \textit{Geographical renaming}, \textit{List of city name changes}, \textit{List of administrative division name changes} as well as lists dedicated to certain countries.

A downside of the format constraint is that we only found lists of geographic entities. This will be ignored for the moment and discussed later in Section~\ref{sec:generalization} where we manually parsed a list of name changes for different kinds of entities and performed the same analysis again. This allows us to argue general applicability of the results to a certain extend.

The parsed lists contain 1,926 distinct entities with 2,852 name changes. For the found names, we fetched a total of 2,782 articles for 1,898 entities. The larger number of articles compared to entities is a result of 766 entities with names that could be resolved to different articles. For 28 entities we were not able to resolve any name to an article.



\begin{table*}[t!]
\setlength{\tabcolsep}{0.5em}
\begin{tabu}{lllX[l]|r|r|r|}
\multicolumn{4}{l|}{\textbf{Subject}} & \textbf{Count} & \multicolumn{2}{c|}{\textbf{Percentage}} \\
\hline
&&&&&&\\[-2ex]
\multicolumn{4}{l|}{Entities} & 1,926 & \textbf{100\%} &\\
& - & \multicolumn{2}{l|}{annotated with change dates} & 708 & 36.8\% &\\
& - & \multicolumn{2}{l|}{resolvable to articles} & 1,898 & 98.5\% & \textbf{100\%}\\
&& - & most current name resolvable & 1,829 & 95.0\% & 96.4\%\\
&& - & linked on a list & 1,786 & 92.7\% & 94.1\%\\
&& - & with multiple articles & 766 & 39.8\% & 40.4\%\\
&& - & annotated with change dates & 696 & 36.1\% & 36.7\%\\
\hline
&&&&&&\\[-2ex]
\multicolumn{4}{l|}{Name changes} & 2,852 & \textbf{100\%} &\\
& - & \multicolumn{2}{l|}{of entities with articles} & 2,810 & 98.5\% &\\
& - & \multicolumn{2}{l|}{annotated with dates} & 933 & 32.7\% &\\
& - & \multicolumn{2}{l|}{of entities with articles, annotated with dates} & 918 & 32.2\% & \textbf{100\%}\\
&& - & mentioned in an article & 572 & 20.1\% & 62.3\%\\
&& - & mentioned in the most current name's article & 551 & 19.3\% & 60.0\%\\
\hline
&&&&&&\\[-2ex]
\multicolumn{4}{l|}{Extracted excerpts} & 572 & \textbf{100\%} &\\
& - & \multicolumn{2}{l|}{sentence distance less than 10} & 488 & 85.3\% & \textbf{100\%}\\
&& - & sentence distance less than 3 & 389 & 68.0\% & 79.7\%\\
&& - & sentence distance 2 & 45 & 7.9\% & 9.2\%\\
&& - & sentence distance 1 & 118 & 20.6\% & 24.2\%\\
&& - & sentence distance 0 & 226 & 39.5\% & 46.3\%\\
\end{tabu}
\vspace{2ex}
\caption{Statistics on name evolutions mentioned on Wikipedia. (percentages are in relation to the 100\% above)}
\label{tab:overview}
\end{table*}

\subsection{Statistics about Name Changes on Wikipedia List Pages}
\label{sec:lists-entity-name}

Table~\ref{tab:overview} shows a detailed listing of the gathered statistics.
The entity with most names in the analysis is \textit{Plovdiv}, the second-largest city of Bulgaria, with 11 changes:

\textit{Kendros (Kendrisos/Kendrisia) $\rightarrow$ Odryssa $\rightarrow$ Eumolpia $\rightarrow$ Philipopolis $\rightarrow$ Trimontium $\rightarrow$ Ulpia $\rightarrow$ Flavia $\rightarrow$ Julia $\rightarrow$ Paldin/Ploudin $\rightarrow$ Poulpoudeva $\rightarrow$ Filibe $\rightarrow$ Plovdiv}

The average number of changes per entity is 1.48. Each of the entities that have changed their name has 2.39 different names in average.

Unlike \textit{Plovdiv}, on 708 (36.8\%) of the entities we extracted at least one name change was annotated with a date. Overall, 933 of the total 2,852 are annotated with dates (32.7\%). Only these were subject of further analysis as the ultimate goal is to identify excerpts that describe a full name change, which we consider consisting of preceding name, succeeding name and change date (s. Section~\ref{sec:procedure}).

An additional requirement in order to extract excerpts is that the entities need to have a corresponding article, which potentially describes the evolution. Out of the 1,926 entities, we found 1,786 entities being linked to an article on one of the lists under consideration. Additionally, we were able to find articles corresponding to names or aliases of 112 entities. In total, 1,898 of the entities could be resolved to Wikipedia articles (98.5\%). These compromise 2,810 name changes, which is also 98.5\% of the initial 2,852. For 1,829 entities the most current name could be resolved to an article (96.4\%). In addition, 114 have a previous name that could be resolved to another article. One reason for this is former names that have a dedicated article, such as \textit{New Amsterdam}\footnote{\url{http://en.wikipedia.org/wiki/New_Amsterdam}}, the former name of New York City.

The intersection between the entities annotated with change dates and the ones with articles is 696, which is 36.1\% of all entities we started with. In terms of name changes, the intersection of name changes with dates and those belonging to entities with articles is 918, which is 32.2\% of the initial name changes. These constitute the subject of our research regarding excerpts describing evolutions.

\subsection{Statistics about Evolution Mentions in Wikipedia Articles}
\label{sec:ment-wikip-articl}

Proceeding with 1,898 entities with articles, we analyzed 2,782 fetched articles. For entities that were resolved to multiple articles, all articles have been taken into account in this analysis. To consider a name change being reported in an article, we checked for all three components of a change to be mentioned (i.e., preceding and succeeding name as well as the change date). This holds for 572 out of the 918 name changes with a date available from the entities with articles (62.3\%). By taking only articles into account that were resolved for the most current name, 551 (60.0\%) complete changes are mentioned.

\begin{figure}[t]
\label{fig:sentence_distance_distribution}
\centering
\includegraphics[width=\columnwidth]{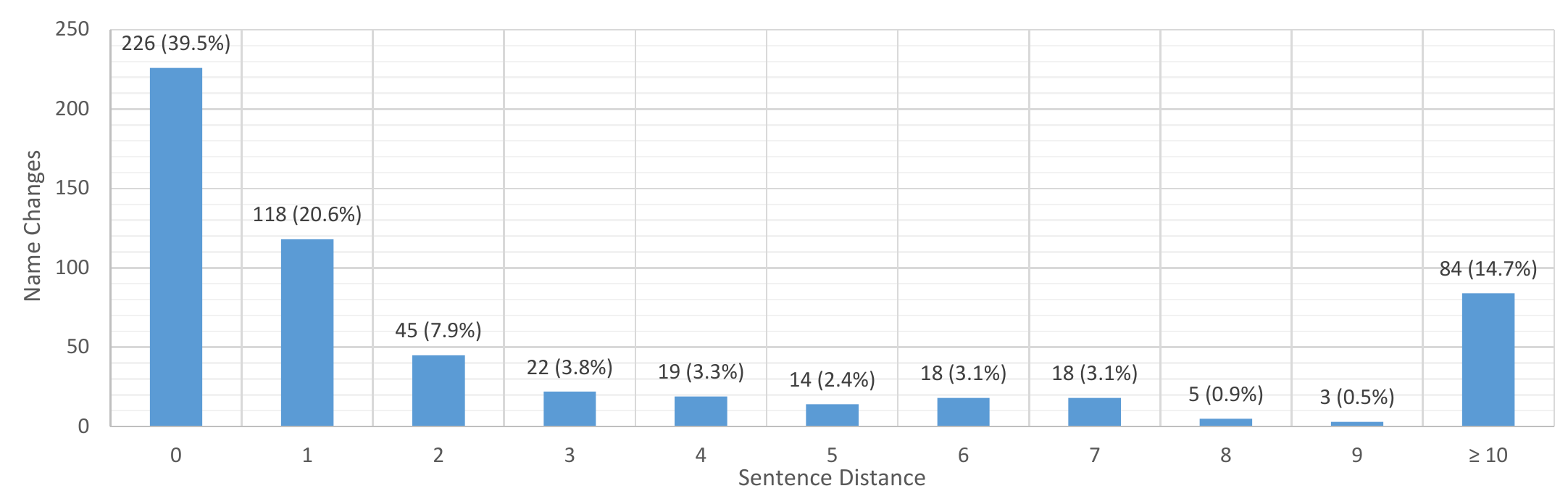}
\caption{Distribution of minimum sentence distances.}
\end{figure}

As described in Section~\ref{sec:research_questions}, we narrowed our objective of the analysis down to the core question: \textit{How many sentences do excerpts span that mention a complete name change?} In order to answer this, we measured the sentence distances of the three components for each of the 572 name changes as described in Section~\ref{sec:procedure}. Overall, the average sentence distance of the extracted excerpts was 19.9. However, this is caused by a very few, very high distances and is not representative, as indicated by the median of 1. In fact, 488 out of the total 572 excerpts, which is 85.3\%, have a distance of less than 10. This shows that for a majority of name changes all three components of a name change occur close together in a text, which positively answers the question if excerpts of limited length cover full name changes (compare Section~\ref{sec:research_questions}). Most likely, those excerpts are dedicated to describe the name evolution of the corresponding entity. As the distribution of sentence distances in Figure~\ref{fig:sentence_distance_distribution} shows, a significant majority of 389 changes, around 80\% (79.7\%) of excerpts with 10 sentences or less, even have a shorter distance of less than three. For our research in terms of extracting evolution patterns, shorter excerpts are more interesting, because they are more likely to constitute excerpts dedicated to describe an evolution. Passages consisting of more than 10 sentences are likely to only cover two independent mentions of preceding and succeeding names of a change. Therefore, in the following we will concentrate on the shortest excerpts with a distance less than three, which constitute the largest amount of all findings. 

\section{Generalization and Discussions}

The found results are supposed to serve as a seed to extract knowledge about name changes of any kind of entities. At this point however, the analyzed lists only consisted of geographic entities. Reliably parsing name changes from different domains on Wikipedia was not possible. In contrast to the semi-structured lists of geographic entities, they are only described in an unstructured way. Thus, we can only carefully make the assumption that our observations hold for entities of other kinds as well. To verify this, we manually parsed a list of renamed products and companies and performed the same analysis in order to generalize the results. In this section we compare and discuss our findings and outline how the results can be used to extend the training set towards general entities. 

\subsection{Towards Generalization}
\label{sec:generalization}

In addition to the automatically parsed lists of different geographic entities, we manually extracted 47 entities from the list of renamed products\footnote{\url{http://en.wikipedia.org/wiki/List_of_renamed_products}}, which also contains company names. For these we performed the same analysis as described in Section~\ref{sec:analysis}. The details of the dataset and statistical results are summarized in Table~\ref{tab:products}. As before, we only proceeded with entities that were resolvable to Wikipedia articles and annotated with change dates on the list. 36 entities meet these conditions, which is 75\%. These consist of 45 (71.4\%) name changes out of the total 63 that are listed on the parsed collection. As for geographic entities, we searched for the succeeding and preceding name as well as the date of the changes in the corresponding articles. We discovered all three components for 36 changes. This corresponds to 80\%, which is even more than the 62.3\% we found for the geographic entities. This result supports our hypothesis that a large amount of name changes are mentioned in the Wikipedia article of the corresponding entity.

\begin{table}[t]
\setlength{\tabcolsep}{0.5em}
\begin{tabu}{lllX[l]|r|r|r|}
\multicolumn{4}{l|}{\textbf{Subject}} & \textbf{Count} & \multicolumn{2}{c|}{\textbf{Percentage}} \\
\hline
&&&&&&\\[-2.5ex]
\multicolumn{4}{l|}{Entities} & 48 & \textbf{100\%} &\\
& - & \multicolumn{2}{l|}{resolvable to articles} & 45 & 93.8\% & \textbf{100\%}\\
&& - & annotated with change dates & 36 & 75.0\% & 80.0\%\\
\hline
&&&&&&\\[-2ex]
\multicolumn{4}{l|}{Name changes} & 63 & \textbf{100\%} &\\
& - & \multicolumn{2}{l|}{of entities with articles} & 59 & 93.7\% &\\
& - & \multicolumn{2}{l|}{annotated with dates} & 48 & 76.2\% &\\
& - & \multicolumn{2}{l|}{of entities with articles, annotated with dates} & 45 & 71.4\% & \textbf{100\%}\\
&& - & mentioned in an article & 36 & 57.1\% & 80.0\%\\
\hline
&&&&&&\\[-2ex]
\multicolumn{4}{l|}{Extracted excerpts} & 36 & \textbf{100\%} &\\
& - & \multicolumn{2}{l|}{sentence distance $<$ 10} & 33 & 91.7\% & \textbf{100\%}\\
&& - & sentence distance less than 3 & 22 & 61.1\% & 66.7\%\\
&& - & sentence distance 2 & 2 & 5.5\% & 6.1\%\\
&& - & sentence distance 1 & 6 & 16.7\% & 18.2\%\\
&& - & sentence distance 0 & 14 & 38.9\% & 42.4\%\\
\end{tabu}
\vspace{2ex}
\caption{Statistics on product name evolutions mentioned on Wikipedia. (percentages are in relation to the 100\% above)}
\label{tab:products}
\end{table}

After extracting the minimum excerpts that cover the found changes we computed the sentence distance of each change as described before. We wanted to know if small excerpts of the articles are dedicated to changes, which we consider true if all three components of the majority of changes are mentioned within the shortest distances. On geographic names, 85.3\% spanned 10 sentences of less. On products and companies this even holds for 91.7\%. Also comparable to our analysis of geographic entities is the fact that the majority of more than two-thirds of these changes are covered by excerpts with three sentences or less. While this includes 79.7\% changes on geographic names, we found 66.7\% on products and companies.

\subsection{Discussion}
\label{sec:discussion}

The analysis was driven by the question about Wikipedia's suitability as a resource for extracting name evolution knowledge. The idea was to extract excerpts from Wikipedia articles that describe name evolutions. This knowledge can be used to build up an entity evolution knowledge base to support other application as well as serve as a ground truth for further research in this field.

Our hypothesis was, that name evolutions are described in short excerpts with a limited length. Based on the statistics, this can be affirmed. Almost 70\% of 918 name changes that were available with corresponding articles and dates are mentioned in the Wikipedia articles of their entities. Out of these, 85.3\% were found within excerpts of ten sentences or less. Two-thirds even had a distance less than three sentences. Regarding the shape of the excerpts, a closer look at some excerpts (s. Listing~\ref{lst:excerpts}) revealed that most of them contain certain words, such as ``became'', ``rename'', ``change''. The extraction of particular patterns in order to train classifiers for the purpose of identifying evolutions automatically remains for future work.

The generalization of the results leads to the conclusion that a knowledge base that uses Wikipedia as a source for name change information would cover at least 41.7\% of all name changes by using only excerpts with a sentence distance of less than three. This is the product of the number of entities with articles (98.5\%), the completely mentioned name changes (62.3\% of the changes of entities with articles and dates annotated) and the number of excerpts with distance less than three (68\%). Certainly, it is not sufficient to use Wikipedia as the only source, but the extracted excerpts constitutes a solid foundation for discovering evolution knowledge on other sources. These could include historical texts, newswire articles and, especially for upcoming changes, social networks and blogs.

In future work we are going to involve more kinds of entities. It is planned to use the geographic name changes as a seed to train classifiers which detect excerpts that follow the same patterns. By setting highest priority on accuracy, we can neglect the fact that our seed only consists of geographic entities. It will definitely lower the recall in the beginning, but as the analysis of products and companies shows, name changes on other kinds are included in Wikipedia in a similar manner. Therefore, we are confident of finding new ones, which will increases our set of name changes for which we can extract new excerpts afterwards. Under the assumption that these changes are mentioned in different ways on Wikipedia and other sources, this approach can incrementally extend the training set. 

\begin{lstlisting}[frame=single,captionpos=b,breaklines=true,breakindent=0pt,float=t,caption={Excerpts covering name changes with sentence distance less than 3.},escapechar=@,label=lst:excerpts,columns=flexible,keepspaces=true]
@\textbf{\textit{--- Nyasaland to Malawi in 1964 (current name: Malawi)}}@
The Federation was dissolved in 1963 and in 1964, Nyasaland gained full independence and was renamed Malawi.

@\textbf{\textit{--- General Emilio Aguinaldo to Bailen in 2012 (current name: Bailen)}}@
The Sangguniang Panglalawigan (Provincial Board) has unanimously approved Committee Report 118-2012 renaming General Emilio Aguinaldo, a municipality in the 7th District of the province, to its original, "Bailen" during the 95th Regular Session.

@\textbf{\textit{--- Western Samoa to Samoa in 1997 (current name: Samoa)}}@ 
In July 1997 the government amended the constitution to change the country's name from Western Samoa to Samoa.

@\textbf{\textit{--- Badajoz to San Agustin in 1957 (current name: San Agustin)}}@
On 20 June 1957, by virtue of Republic Act No. 1660, the town's name of Badajoz was changed to San Agustin.

@\textbf{\textit{--- Bombay to Mumbai in 1996 (current name: Mumbai)}}@
In 1960, following the Samyukta Maharashtra movement, a new state of Maharashtra was created with Bombay as the capital. The city was renamed Mumbai in 1996.
\end{lstlisting}

\section{Related Work}

Most related to the long-term aim of this work, a knowledge base dedicated to entity evolution, is YAGO2 \cite{YAGO2}. It is an endeavor to extend the original YAGO knowledge base with temporal as well as spatial information. Most relevant to us is the temporal data, which YAGO2 incorporates to enhance entities as well as facts. In contrast to our aim, they do not gather this data by extracting new knowledge. Instead, they use temporal information which has already been extracted for YAGO and connect it to the corresponding entity or fact. For instance, date of birth and date of death are considered as a person's time of existence. Therefore, dates of name changes are still missing as they are not present in YAGO either.

In terms of the process and analysis results presented in this paper, the related work can be divided into two areas: related research on Wikipedia as well as on entity evolution.

\subsection*{Research on Wikipedia}
\label{sec:research-wikipedia}

A prominent research topic in the context of Wikipedia is prediction of quality flaws. It denotes the task of automatically detecting flaws according to Wikipedia's guidelines, something not to neglect when working with Wikipedia. Anderka et al. \citep{AnderkaSIGIR12} have done an impressive work in this field and give a nice overview of the first challenge dedicated to this topic \cite{AnderkaCLEF12}. Another related topic is the research on Wikipedia's revision history and talk pages. This could also serve as an additional resource for name evolutions in the future. \citet{Ferschke2012} are working on automatically annotating discussions on talk pages and eventually link these to the corresponding content on Wikipedia articles. Additionally, they provide a toolkit for accessing Wikipedia's history \cite{FerschkeACL11}.

Besides research on Wikipedia's infrastructure, many analyses on Wikipedia data have been done. Recently \citet{Goldfarb2012} analyzed the temporal dimension of links on Wikipedia, i.e., the time distance a link bridges when connecting artists from different eras. However, to the best of our knowledge, no analysis has tackled the question on named entity evolutions in Wikipedia articles before.

\subsection*{Research on Entity Evolution}
\label{sec:rese-entity-evol}

Most of the prominent research in the field of entity evolution focuses on query translation. \citet{Berberich2009} proposed a  query reformulation technique to translate terms used in a query into terms used in older texts by connecting terms through their co-occurrence context today and in the past. \citet{Kaluarachchi2010} proposed another approach for computing temporally and semantically related terms using machine learning techniques on verbs shared among them. Similarly to our approach, \citet{Kanhabua2010} incorporate Wikipedia for detecting former names of entities. However, they exploit the history of Wikipedia and consider anchor texts at different times pointing to the same entity as time-based synonyms. This is a reasonable approach, however, as we showed, Wikipedia is not complete in terms of name evolutions. As anchor texts can occur in arbitrary contexts, they are not suitable for pattern discovery as we proposed and thus, cannot be used to extend the knowledge by incorporating other sources, either. Not focused on name changes, but the evolution of an entity's context, presented \citet{Mazeika2011} a tool for visually analyzing entities by means of timelines that show co-occurring entities. Tahmasebi et al. \citep{Tahmasebi2012, Tahmasebi2013} proposed an unsupervised method for named entity evolution recognition in a high quality newspaper (i.e., New York Times). Similarly to the excerpts that we extracted from Wikipedia, they also aim for texts describing a name change. Instead of taking rules or patterns into account, they consider all co-occurring entities in change periods as potential succeeding names and filter them afterwards using different techniques. Based on that method the search engine \textit{fokas}\cite{fokas} has been developed to demonstrate how awareness of name evolutions can support information retrieval. We adapted the method to work on Web data, especially on blogs \cite{Holzmann2013}. However, big drawbacks of this approach are that change periods need to be known or detected upfront and the filters do not incorporate characteristics of the texts indicating a name change.

\section{Conclusions and Future Work}
\label{sec:conclusions}

In our study we investigated how name changes are mentioned in Wikipedia articles regardless of structural elements and found that a large majority is covered by short text passages. Using lists of name changes, we were able to automatically extract the corresponding excerpts from articles. Although the name evolutions mentioned in Wikipedia articles by far cannot be called complete, they provide a respectable basis for discovering more entity evolutions. In future work, we are going to use the excerpts that we found on Wikipedia for discovering patterns and training classifiers to find similar excerpts on other sources as well as unconsidered articles on Wikipedia. The first step on this will be a more detailed analysis of the extracted excerpts, followed by engineering appropriate features. As soon as we are able to identify the components of a name change in the excerpts, we can also incorporate other language versions of Wikipedia. In the long run, we are going to build a knowledge base dedicated to entity evolutions. Such a knowledge base can serves as a source for application that rely on evolution knowledge, like information retrieval systems, especially on Web archives. Furthermore, it constitutes a ground truth for future research in the field of entity evolution, like novel algorithms for detecting entity evolutions on Web content streams.


\begingroup
\let\clearpage\relax
\vspace{3em}
\bibliography{bib}{}
\bibliographystyle{unsrtnat}
\endgroup
\end{document}